# Comparison of Genetic Algorithm and Simulated Annealing Technique for Optimal Path Selection In Network Routing

T.R.Gopalakrishnan Nair
Director – Research Center (RIIC)
Dayananda Sagar Institutions,
Bangalore-78, India.
trgnair@ieee.org

Kavitha Sooda
Associate Member RIIC, DSI Assistant Professor, Dept. of CSE, NMIT
Bangalore-64, India.
kavithasooda@gmail.com

*Abstract*— **This paper addresses the path selection problem from a known sender to the receiver. The proposed work shows path selection using genetic algorithm (GA) and simulated annealing (SA) approaches. In genetic algorithm approach, the multi point crossover and mutation helps in determining the optimal path and also alternate path if required. The input to both the algorithms is a learnt module which is a part of the cognitive router that takes care of four QoS parameters. The aim of the approach is to maximize the bandwidth along the forward channels and minimize the route length. The population size is considered as the N nodes participating in the network scenario, which will be limited to a known size of topology. The simulated results show that, by using genetic algorithm approach, the probability of shortest path convergence is higher as the number of iteration goes up whereas in simulated annealing the number of iterations had no influence to attain better results as it acts on random principle of selection.**

***Keywords-routing; cognition; genetic algorithm; multipoint cross over; simulated annealing;bandwidth; QoS.***

## I. INTRODUCTION

Routing is a process of forwarding the data from a known sender to the receiver. In this processing, the data may travel through several intermediate paths, where there is a need to select the best possible optimal nodes to forward the data. This selection of nodes will enable to achieve a high performance in the network. There are many existing work done in this area of path selection which are discussed in the literature [1]-[8]. The existing algorithms have found the optimal paths considering either one or two QoS parameters or hop counts or cost as the deciding factor for route selection. The proposed work assumes four QoS parameters such as bandwidth, delay, jitter and loss which act as the input to both GA and SA approaches and bandwidth availability at the links for finding the optimal path.

In order to make the present network systems to be intelligent there is a need for an open platform for cognitive experiments. A common building block has been proposed in [9]. Bandwidth availability has been determined by multi hop analysis [10]. Setting up the geographical layout for cognitive networks is described in [11]. A model which combines a reconfigurable core and control systems along with genetic algorithms for cognitive functionality has been dealt in [12]. The security aspects are dealt in [13] which discuss the research challenges for security in cognitive networks. Among the major key security aspects which are dealt in [13], the communication control channel jamming congestion is automatically avoided by our approach as the data is forwarded based on the availability of bandwidth at the given link. The proposed work follows the architecture model as in [14] where as the deciding module has been dealt with reasoning capability.

The intention objective of this paper is to find an efficient solution for end-to-end delivery [15, 16] which involves geographical intelligence and multiple router integration at large distance. However we handle several layers of routers to prove GA and SA based selection of channel which can be used in cognitive routing.

Section 2 addresses the fitness function, crossover, mutation methods and simulated annealing. The simulation results are shown in Section 3, and the conclusions and future works are dealt with in section 4.

## II. GENETIC ALGORITHM AND SIMULATED ANNEALING ALGORITHM

Genetic algorithms are a part of evolutionary computing. It is also an efficient search method that has been used for path selection in networks. GA is a stochastic search algorithm which is based on the principle of natural selection and recombination. A GA is composed with a set of solutions, which represents the chromosomes. This composed set is referred to population. Population consists of set of chromosome which is assumed to give solutions. From this population, we randomly choose the first generation from which solutions are obtained. These solutions become a part of the next generation. Within the population, the chromosomes are tested to see whether they give a valid solution. This testing operation is nothing but the fitness functions which are applied on the chromosome. Operations like selection, crossover and mutation are applied on the selected chromosome to obtain the progeny. Again fitness function is applied to these progeny to test for its fitness. Most fit progeny chromosome will be the participants in the next generation. The best sets of solution are obtained using heuristic search techniques. The general description of GA is as follows:

a. **First Generation** randomly pick n chromosome to form a population assuming that this could be the probable solution to the problem.

b. **Fitness Function** the fitness function f(x) is applied on each chromosome in the generation.

c. **Next Generation** create the next generation by performing the following steps until n chromosomes are obtained

   i. **Selection operation** Select any two best fittest chromosome from the generation

   ii. **Crossover** with a defined probability apply the crossover technique for the above obtained chromosome to form the children

   iii. **Mutation** with a defined probability mutate a new gene at desired position,

d. **Test** whether the obtained children are fit to go to next generation. If yes, then move them to next generation.

e. **Test** if generation is of desired size, if yes, stop, and return the best solution from the current generation.

f. **Repeat** go back to b

The performance of GA is based on efficient representation, evaluation of fitness function and other parameters like size of population, rate of crossover, mutation and the strength of selection. Genetic algorithms are able to find out optimal or near optimal solution depending on the selection function [17, 18].

Simulated annealing (SA) algorithm [20-21] is a general purpose optimization technique. It has been derived from the concept of metallurgy is which we have to crystallize the liquid to required temperature. In this process the liquids will be initially at high temperature and the molecules are free to move. As the temperature goes down, there shall be restriction in the movement of the molecules and the liquid begins to solidify. If the liquid is cooled slowly enough, then it forms a crystallize structure. This structure will be in minimum energy state. If the liquid is cooled down rapidly then it forms a solid which will not be in minimum energy state. Thus the main idea in simulated annealing is to cool the liquid in a control matter and then to rearrange the molecules if the desired output is not obtained. This rearrangement of molecules will take place based on the objective function which evaluates the energy of the molecules in the corresponding iterative algorithm. SA aims to achieve global optimum by slowly converging to a final solution, making downwards move hoping to reach global optimum solution. Given a solution $S_s$ we select the neighbor solution $S_n$ and the difference is calculated using the objective function,

$$\Delta f = f(S_n) - f(S_s) \quad (1)$$

If the function improves the value, i.e, if ($\Delta f < 0$),then replace the current solution with the new one. If ($\Delta f \geq 0$) , then the new solution is accepted with a probability factor of $p(\Delta f) = \exp(-\Delta f/T)$, where T is the temperature which is the controlling parameter. The procedure is repeated until the terminating condition is met which is as follows

- For a considered temperature, repeat for certain steps
- The considered temperature must be greater than minimum temperature

### *A. Proposed Algorithm*

The proposed algorithm follows the above mentioned steps to obtain the optimal path. The input to the path selection GA and SA is the set of nodes which would satisfy the four QoS parameters (bandwidth, delay, jitter and loss). This kind of selection is made possible by sending information in the packet of the node itself [19]. This kind of selection would lead to almost an optimal solution with respect to QoS. From this set of nodes, we would calculate the available bandwidth ($A_b$) using the formula,

$A_{b^{m}_{j=1}}$ = link utility- required bandwidth by the data to be sent.

$$A_{b_{j=1}}^{m} = Link_{utility} - Required_{Bythenode} \qquad (2)$$

The link utility is stored in a vector and referred for calculation in $A_b$. If $A_b>0$, then the link can participate in the optimal path, otherwise it is not chosen. The fitness function for GA/ objective function for SA are calculated using,

Path selection fj (t) =Available bandwidth/ summation of $i^{th}$ chromosome.

$$f_j(t) = \frac{Ab_j}{\sum_{i=1}^{m} A_{b_i}(t)} \qquad (3)$$

If $f_j(t)$ is between 0.5 and 1 those chromosomes gets selected and data is forwarded in that path after the convergence of the generations.

*a. The algorithm for GA is as follows:*

**begin** PATHSELECTION_GA

Create initial population of n nodes randomly.

**while** generation_count < k do

/* k = max. no. of generations.*/

**begin**

Selection

Fitness Function

Modified crossover

Mutation

Increment generation_count.

**end** ;

Output the optimal path by selecting the highest probability value chromosome on which data can be sent

**end** PATHSELECTION_GA.

*b. The algorithm for SA is as follows:*

**begin** PATHSELECTION_SA

Initialize the T, Tstop, ts, α, N.

Choose **Ss**

Assign tstop=ts

**while** tstop > 0 and T>Tstop **do**

**begin**

**for** i=1 to N **do**

**begin**

generate new solution at random, $S_n$

calculate $\Delta f=f(S_n) - f(S_s)$

if $(\Delta f < 0)$ then

$S_s = S_n$

else

if (exp(-Δf/T) > rand( 0,1) )

$S_s = S_n$

if $S_n$ is chosen then

tstop=ts

else

tstop--

**end ;**

T=T * α

**end;**

Output the optimal path by selecting the $S_c$ which would be recorded after every change on it.

**end** PATHSELECTION_SA

## *B. Representation*

The network under consideration is represented as G = (V, E), a connected non-loop free graph with N nodes. The metric of optimization is bandwidth available between the nodes. The goal is to find the path with availability of bandwidth between source node $V_s$ and destination $V_d$, where $V_s$ and $V_d$ belong to V. E is the set of edges connecting the nodes which are represented in V. From this topology we develop a graphical representation of QoS satisfied by the nodes and generation of optimal path using genetic algorithm. Finally data is sent along the generated path.

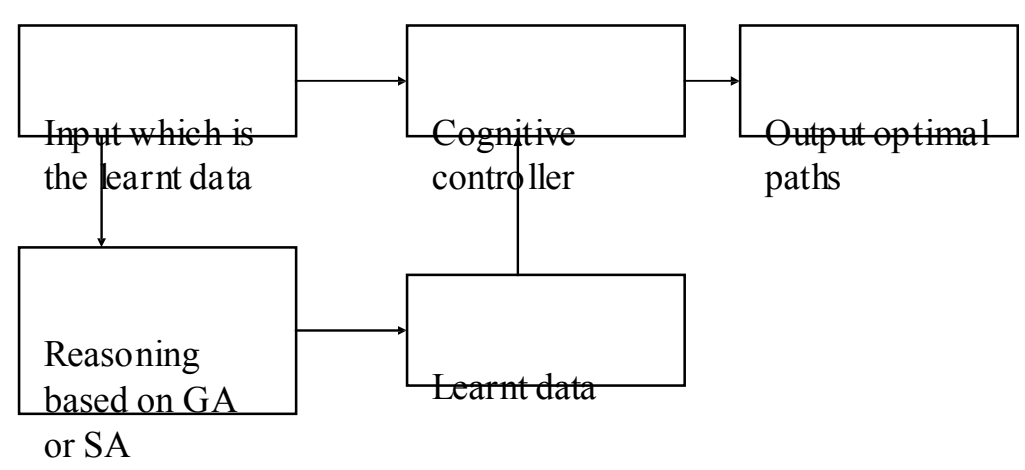

Figure 1. Cognition Model

### C. Assumptions for GA

GA uses a selection mechanism to select the individuals from the population to insert into the mating pool. Individuals from the mating pool are used to generate new off spring which will participate in the next generation. As the individuals of the next generation are going to participate further this is better for the genes to be of good condition. This selection function leads to a better population with good condition. Here the selection process is carried out by Roulette Wheel method. In this method, the individuals are chosen based on the relative fitness with its competitors. A reference chromosome starting at source and ending at destination will always be selected in every population by elitism.

In GA, crossover operator combines sub parts of two parent chromosomes and produces off spring that contains some parts of both the parent. Here we consider both single point and multipoint crossover technique. In single point crossover technique, one offspring consists of first part of one parent and second part of the other parent. Similarly the other offspring is generated. Here we also use multi point crossover mechanism called partially mapped crossover. In this two chromosomes are picked at random. The strings between the crossover sites are exchanged position by position; other elements are determined by ordering information, which is partially determined by each of its parents.

Sometimes it may be possible that by crossover operation, a new population never gets generated. To overcome this limitation, we do mutation operation. Here we use insertion method, as a node along the optimal path may be eliminated through crossover.

The probability of selecting the chromosome $f_j(t)$, is given by,

$$P_j(t) = \frac{f_j(t)}{\sum_{i=1}^{m} f_j(t)} \qquad (4)$$

### D. Assumption for SA Algorithm

The start temperature and the end temperature are chosen at random. The algorithm is performed on a given temperature N times. The stopping condition ts, is also chosen along with reducing cooling parameter, α. The objective function is same as the fitness function taken which is applied in GA.

## III. SIMULATION AND RESULTS

Current work is tested on the network consisting of ten nodes. The topology of the network is shown in Fig. 2.

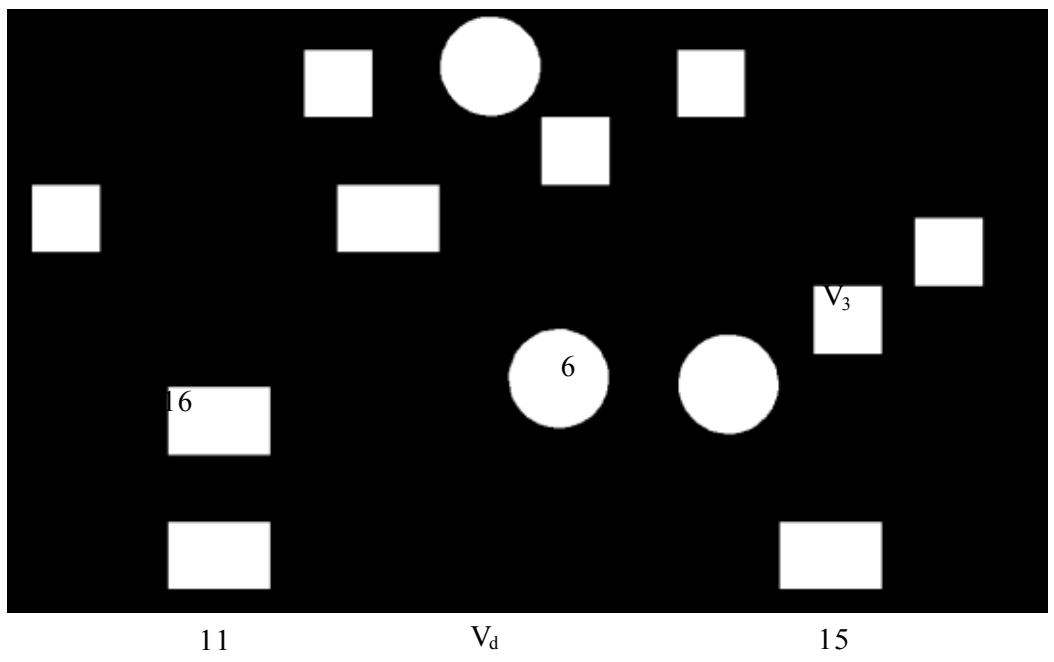

Figure 2: QoS Selection Map

Here the paper distinguishes itself by deciding the channels based on bandwidth availability. The result obtained is based on the best bandwidth available at minimum hop count.

For GA, initially ten random chromosomes are generated and placed in the roulette wheel based on the path length. Shorter the path length higher is the probability of selecting the chromosome from the roulette wheel. Out of the ten random chromosomes generated five best are considered for the first generation. Here the result is shown for three such generations which are obtained by applying crossover and insertion mutation functions. At each generation of population, validation of the chromosome is carried out and the best fit chromosome is only considered for next generation. It was found that probability of shortest path convergence was faster. The results obtained are shown in the table for the three generations.

The first generation input was chosen randomly based on roulette wheel selection method, for which fitness function is applied and probability of its survival has been calculated as shown below:

TABLE I. GENERATION 1

| Fitness | Chromosomes | No. of nodes visited | Probability of selecting chromosome |
|---|---|---|---|
| 0.2903 | C1 | 3 | 1 |
| 0 | C4 | 4 | 0 |
| 0.3030 | C3 | 3 | 0.5106 |
| 0.1766 | C8 | 5 | 0.7699 |
| 0 | C7 | 5 | 0 |

From the first generation we see that crossover probability on any two chromosome of equal length does not produce any children. So insertion mutation was applied on C3 chromosome to obtain C6 chromosome. By elitism C1 chromosome exist and remaining are obtained by the selection method. The results are as in table 2.

TABLE II. GENERATION 2

| Fitness | Chromosomes | No. of nodes visited | Probability of selecting chromosome |
|---|---|---|---|
| 0.2903 | C1 | 3 | 1 |
| 0.2045 | C5 | 4 | 0.4132 |
| 0.2173 | C6 | 4 | 0.3052 |
| 0.1267 | C9 | 6 | 0.1510 |
| 0 | C10 | 6 | 0 |

The third generation children are obtained from the multipoint crossover technique on C9 and C10 chromosomes to obtain C11. The remaining chromosomes are obtained as in second generation. Similarly the fourth generation was obtained by single point crossover on C1 and C3 to obtain C12. The fifth generation chromosome did not produce any valid paths after crossover technique and since it was the last generation and mutation operation was to be applied only 0.01 percent of time, the chromosome was selected based on selection method. The results are shown in the tables below.

TABLE III. GENERATION 3

| Fitness | Chromosomes | No. of nodes visited | Probability of selecting chromosome |
|---|---|---|---|
| 0.2903 | C1 | 3 | 1 |
| 0 | C11 | 6 | 0 |
| 0.1267 | C9 | 6 | 0.3038 |
| 0.3030 | C3 | 3 | 0.4208 |
| 0 | C7 | 5 | 0 |

From the result obtain, we apply the minimum path algorithm to obtain the optimal path. The criteria of this algorithm are to see that it has to traverse through less number of nodes and must posses higher probability value. Thus we find that path length of three which is the shortest path, having higher probability, has been selected for optimal path selection.

For SA, the following are the initial values considered for simulation:

TABLE IV. INITIAL ASSUMPTIONS FOR SA

| T | Tstop | N | α | ts |
|---|---|---|---|---|
| 1000 | 50 | 2 | 1/5 | 5 |

The following table shows the result obtained for the above assumption:

TABLE V. RESULTS OBTAINED FROM SA

| $S_c$ | $S_n$ | Δf | Tstop | T |
|---|---|---|---|---|
| C1 | C3 | -0.0127 | 5 | 1000 |
| C3 | C6 | -0.0857 | 4 | |
| C3 | C8 | -0.1264 | 3 | 200 |
| C3 | C9 | -0.1703 | 2 | |
| C3 | | | | 40 |

From the above result we see that the convergence to obtain shortest path has been quick for the initial random selection of the path. But it is not always true if at random longest path was selected. Therefore it is necessary to record the output obtained for every T. Once the terminating condition is met, we need to select the best $S_c$ by choosing minimum value of Δf, in order to ensure the shortest path.

For the results obtained by applying GA and SA we see that C3 has been selected as the best path of length 3, i.e., Vs-V3-V7-Vd.

## IV. CONCLUSION AND FUTUTE WORK

This work presented an comparative study for finding optimal path selection technique using genetic algorithm and simulated annealing algorithm. Here the network performance was maximized using the four QoS parameters. Here the data was forwarded based on the bandwidth availability. The results show better convergence of shortest length chromosome using GA than SA.

Here the best selection of the path was obtained only based on shortest hop count, which does not have the geographical implications.